\documentclass[10pt,twocolumn,letterpaper]{article}

\usepackage[usenames,dvipsnames]{color}
\usepackage{cvpr,times,epsfig,graphicx,amsmath,amssymb}
\usepackage{algorithm,algorithmic,booktabs}
\usepackage[pagebackref=true,breaklinks=true,letterpaper=true,colorlinks,bookmarks=false]{hyperref}

\cvprfinalcopy
\def\cvprPaperID{866}

\ifcvprfinal\fi

\newcommand{\vocab}{\mathcal{V}}
\newcommand{\vocabThres}{\tilde{\mathcal{V}}}
\newcommand{\sosTok}{\scalebox{0.85}{\ensuremath{<\!\!s\!\!>}}}
\newcommand{\eosTok}{\scalebox{0.85}{\ensuremath{<\!\!/s\!\!>}}}
\newcommand{\insertH}[2]{\includegraphics[height=#1\textwidth]{#2}}
\newcommand{\insertW}[2]{\includegraphics[width=#1\textwidth]{#2}}

\begin{document}

\title{From Captions to Visual Concepts and Back}

\author{
\begin{tabular}{c@{\hspace{8mm}}c@{\hspace{8mm}}c@{\hspace{8mm}}c}
 Hao Fang\thanks{H. Fang, S. Gupta, F. Iandola and R. K. Srivastava contributed equally to this work while doing internships at Microsoft Research. Current affiliations are H. Fang: University of Washington; S. Gupta and F. Iandola: University of California at Berkeley; R. K. Srivastava: IDSIA, USI-SUPSI.} & Saurabh Gupta\footnotemark[1] & Forrest Iandola\footnotemark[1] & Rupesh K. Srivastava\footnotemark[1] \\
 Li Deng & Piotr Doll{\'a}r{\thanks{P. Doll\'ar is currently at Facebook AI Research.}} & Jianfeng Gao & Xiaodong He \\
 Margaret Mitchell & John C. Platt{\thanks{J. Platt is currently at Google.}}  & C. Lawrence Zitnick & Geoffrey Zweig
\end{tabular}\\ \\
{\Large Microsoft Research} }

\maketitle

\begin{abstract}
This paper presents a novel approach for automatically generating image descriptions: visual detectors, language models, and multimodal similarity models learnt directly from a dataset of image captions. We use multiple instance learning to train visual detectors for words that commonly occur in captions, including many different parts of speech such as nouns, verbs, and adjectives. The word detector outputs serve as conditional inputs to a maximum-entropy language model. The language model learns from a set of over 400,000 image descriptions to capture the statistics of word usage. We capture global semantics by re-ranking caption candidates using sentence-level features and a deep multimodal similarity model. Our system is state-of-the-art on the official Microsoft COCO benchmark, producing a BLEU-4 score of 29.1\%. When human judges compare the system captions to ones written by other people on our held-out test set, the system captions have equal or better quality 34\% of the time.
\end{abstract}

\section{Introduction}\label{sec:introduction}

When does a machine ``understand'' an image? One definition is when it can generate a novel caption that summarizes the salient content within an image. This content may include objects that are present, their attributes, or their relations with each other. Determining the salient content requires not only knowing the contents of an image, but also deducing which aspects of the scene may be interesting or novel through commonsense knowledge \cite{zitnickCVPR13,ChenICCV2013neil,levan}.

\begin{figure}\centering
 \includegraphics[width=0.45\textwidth]{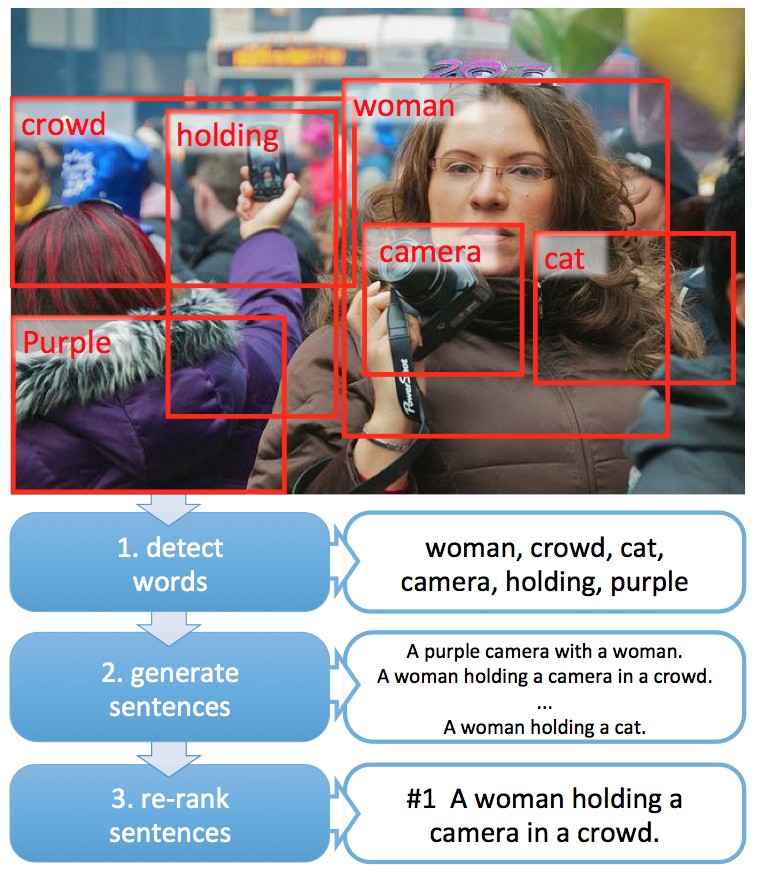}
 \caption{An illustrative example of our pipeline.}\label{fig:teaser}
\end{figure}

This paper describes a novel approach for generating image captions from samples. We train our caption generator from a dataset of images and corresponding image descriptions. Previous approaches to generating image captions relied on object, attribute, and relation detectors learned from separate hand-labeled training data \cite{yang2011corpus,kulkarni2011baby}.

The direct use of captions in training has three distinct advantages. First, captions only contain information that is inherently salient. For example, a dog detector trained from images with captions containing the word \texttt{dog} will be biased towards detecting dogs that are salient and not those that are in the background. Image descriptions also contain variety of word types, including nouns, verbs, and adjectives.  As a result, we can learn detectors for a wide variety of concepts. While some concepts, such as \texttt{riding} or \texttt{beautiful}, may be difficult to learn in the abstract, these terms may be highly correlated to specific visual patterns (such as a person on a horse or mountains at sunset).

Second, training a language model (LM) on image captions captures commonsense knowledge about a scene. A language model can learn that a person is more likely to sit on a chair than to stand on it. This information disambiguates noisy visual detections.

Third, by learning a joint multimodal representation on images and their captions, we are able to measure the global similarity between images and text, and select the most suitable description for the image.

An overview of our approach is shown in Figure \ref{fig:teaser}. First, we use weakly-supervised learning to create detectors for a set of words commonly found in image captions. Learning directly from image captions is difficult, because the system does not have access to supervisory signals, such as object bounding boxes, that are found in other data sets \cite{PASCAL, imagenet_cvpr09}. Many words, e.g., \texttt{crowded} or \texttt{inside}, do not even have well-defined bounding boxes. To overcome this difficulty, we use three ideas. First, the system reasons with image sub-regions rather than with the full image. Next, we featurize each of these regions using rich convolutional neural network (CNN) features, fine-tuned on our training data \cite{krizhevskyNIPS12,simonyan14very}. Finally, we map the features of each region to  words likely to be contained in the caption. We train this map using multiple instance learning (MIL) \cite{maron98,zhangNIPS05} which learns discriminative visual signature for each word.

Generating novel image descriptions from a bag of likely words requires an effective LM. In this paper, we view caption generation as an optimization problem. In this view, the core task is to take the set of word detection scores, and find the highest likelihood sentence that covers each word exactly once. We train a maximum entropy (ME) LM from a set of training image descriptions \cite{Berger1996,ratnaparkhi2002trainable}. This training captures commonsense knowledge about the world through language statistics \cite{NELL}. An explicit search over word sequences is effective at finding high-likelihood sentences.

The final stage of the system (Figure \ref{fig:teaser}) re-ranks a set of high-likelihood sentences by a linear weighting of sentence features. These weights are learned using Minimum Error Rate Training (MERT) \cite{Och2003}. In addition to several common sentence features, we introduce a new feature based on a Deep Multimodal Similarity Model (DMSM). The DMSM learns two neural networks that map images and text fragments to a common vector representation in which the similarity between sentences and images can be easily measured. As we demonstrate, the use of the DMSM significantly improves the selection of quality sentences.

To evaluate the quality of our automatic captions, we use three easily computable metrics and {\it better/worse/equal} comparisons by human subjects on Amazon's Mechanical Turk (AMT). The evaluation was performed on the challenging Microsoft COCO dataset \cite{linECCV14,capeval2015} containing complex images with multiple objects. Each of the 82,783 training images has 5 human annotated captions. For measuring the quality of our sentences we use the popular BLEU \cite{papineni2002bleu}, METEOR \cite{banerjee2005meteor} and perplexity (PPLX) metrics. Surprisingly, we find our generated captions outperform humans based on the BLEU metric; and this effect holds when evaluated on unseen test data from the COCO dataset evaluation server, reaching 29.1\% BLEU-4 vs.~21.7\% for humans. Human evaluation on our held-out test set has our captions judged to be of the same quality or better than humans 34\% of the time. We also compare to previous work on the PASCAL sentence dataset \cite{Rashtchian2010}, and show marked improvements over previous work. Our results demonstrate the utility of training both visual detectors and LMs directly on image captions, as well as using a global multimodal semantic model for re-ranking the caption candidates.

\section{Related Work}\label{sec:related}

There are two well-studied approaches to automatic image captioning: retrieval of existing human-written captions, and generation of novel captions. Recent retrieval-based approaches have used neural networks to map images and text into a common vector representation \cite{socher2013grounded}. Other retrieval based methods use similarity metrics that take pre-defined image features \cite{hodosh2013framing,ordonez2011im2text}. Farhadi et al. \cite{farhadi2010every} represent both images and text as linguistically-motivated semantic triples, and compute similarity in that space. A similar fine-grained analysis of sentences and images has been done for retrieval in the context of neural networks \cite{karpathy2014deep}.

Retrieval-based methods always return well-formed human-written captions, but these captions may not be able to describe new combinations of objects or novel scenes. This limitation has motivated a large body of work on generative approaches, where the image is first analyzed and objects are detected, and then a novel caption is generated. Previous work utilizes syntactic and semantic constraints in the generation process \cite{mitchell2012midge,yao2010i2t,li2011composing,kuznetsova2012collective,kulkarni2011baby,yang2011corpus}, and we compare against prior state of the art in this line of work.  We focus on the Midge system \cite{mitchell2012midge}, which combines syntactic structures using maximum likelihood estimation to generate novel sentences; and compare qualitatively against the Baby Talk system \cite{kulkarni2011baby}, which generates descriptions by filling sentence template slots with words selected from a conditional random field that predicts the most likely image labeling.  Both of these previous systems use the same set of test sentences, making direct comparison possible.

\begin{figure}\centering
    \includegraphics[height=0.2185\textwidth,width=0.23\textwidth]{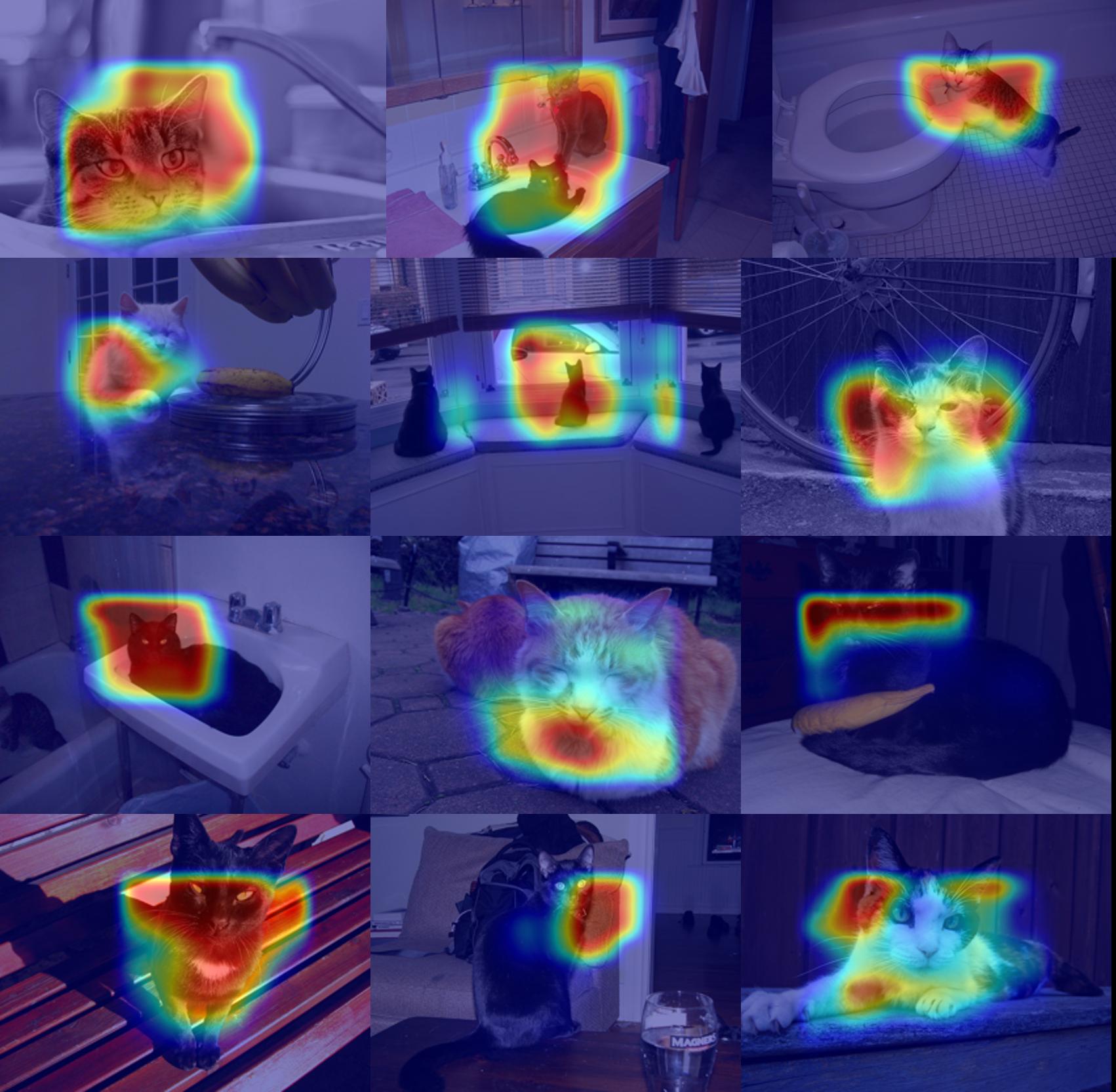}
    \insertW{0.23}{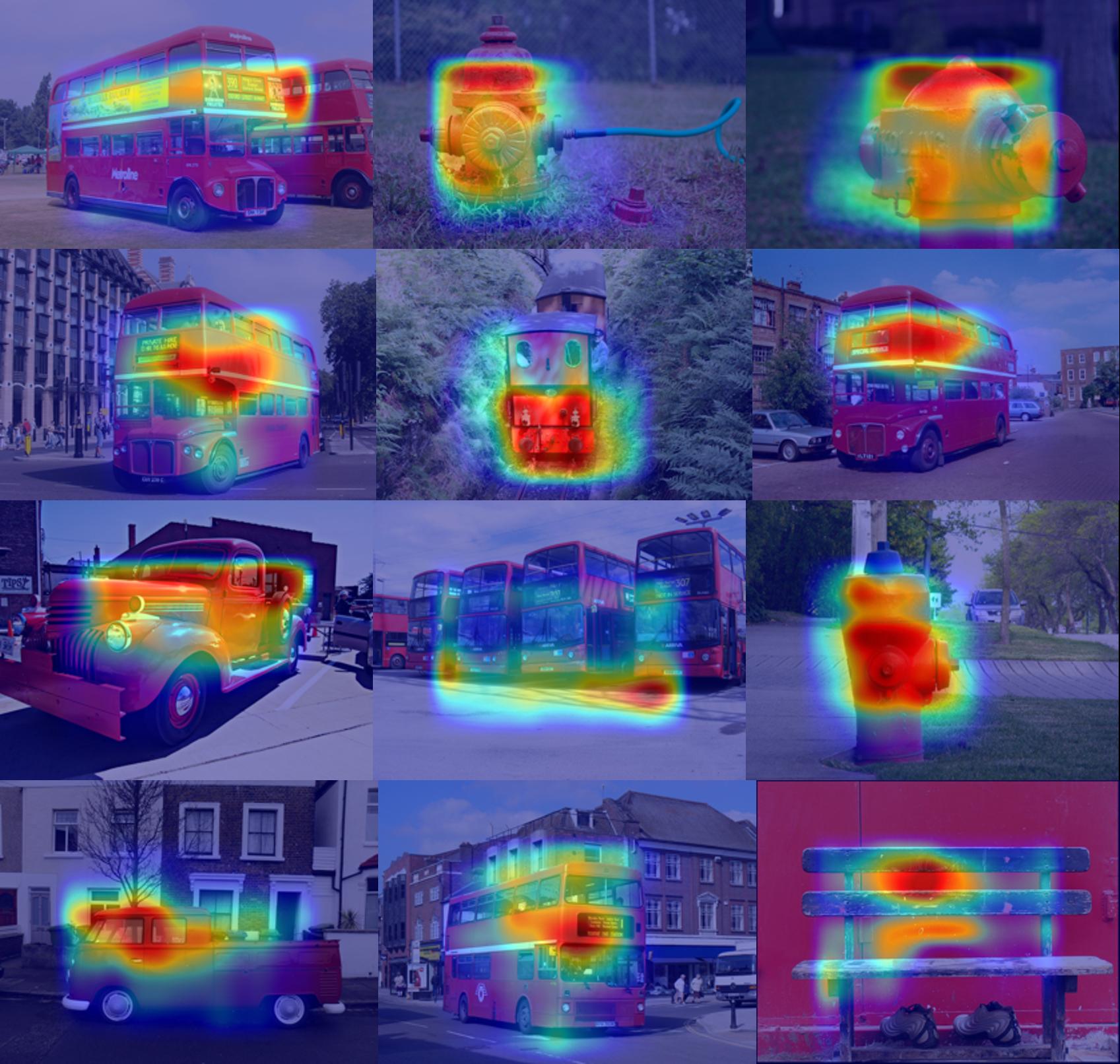}\\
    \vspace{1pt}
    \includegraphics[height=0.209\textwidth,width=0.23\textwidth]{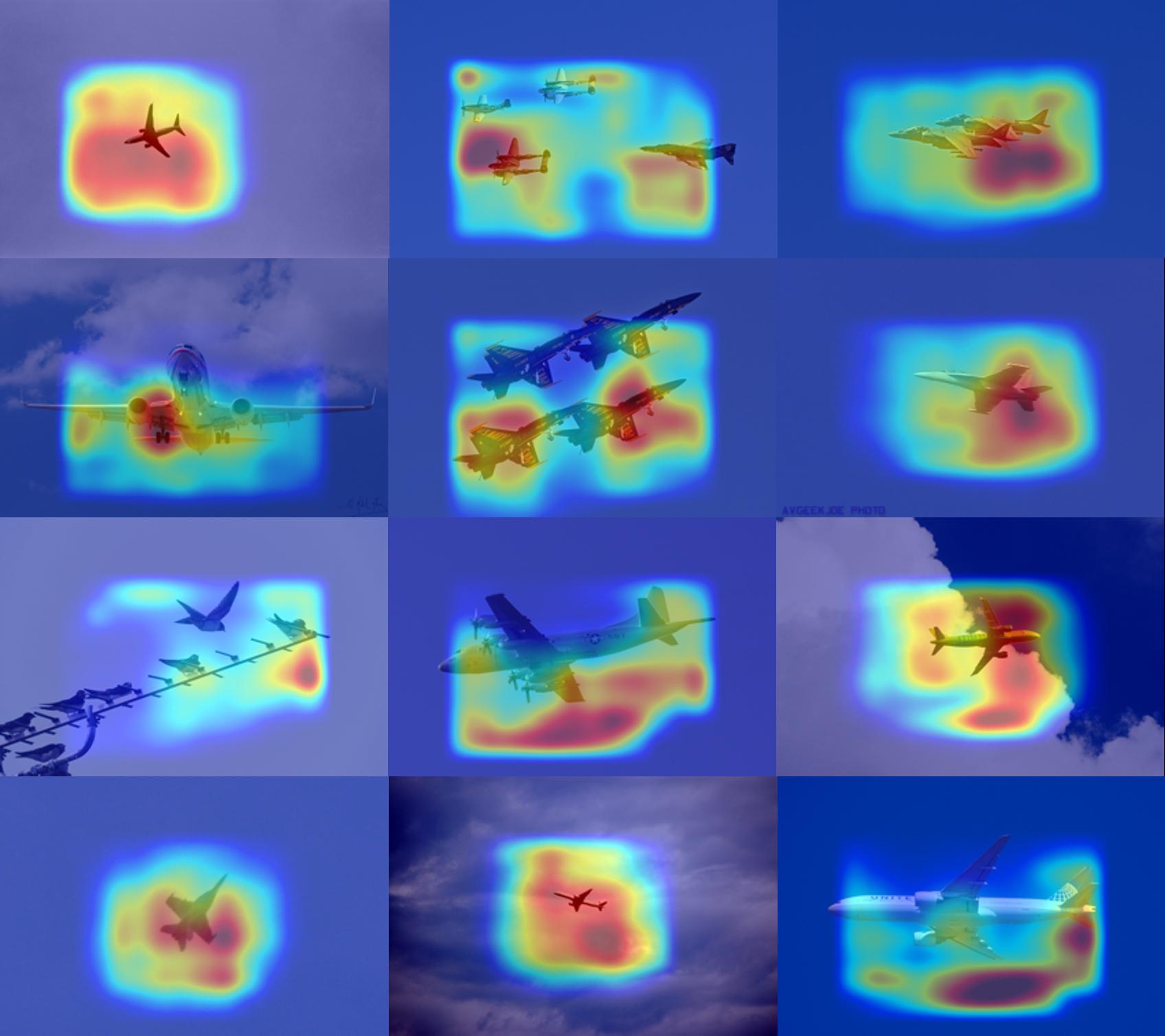}
    \insertW{0.23}{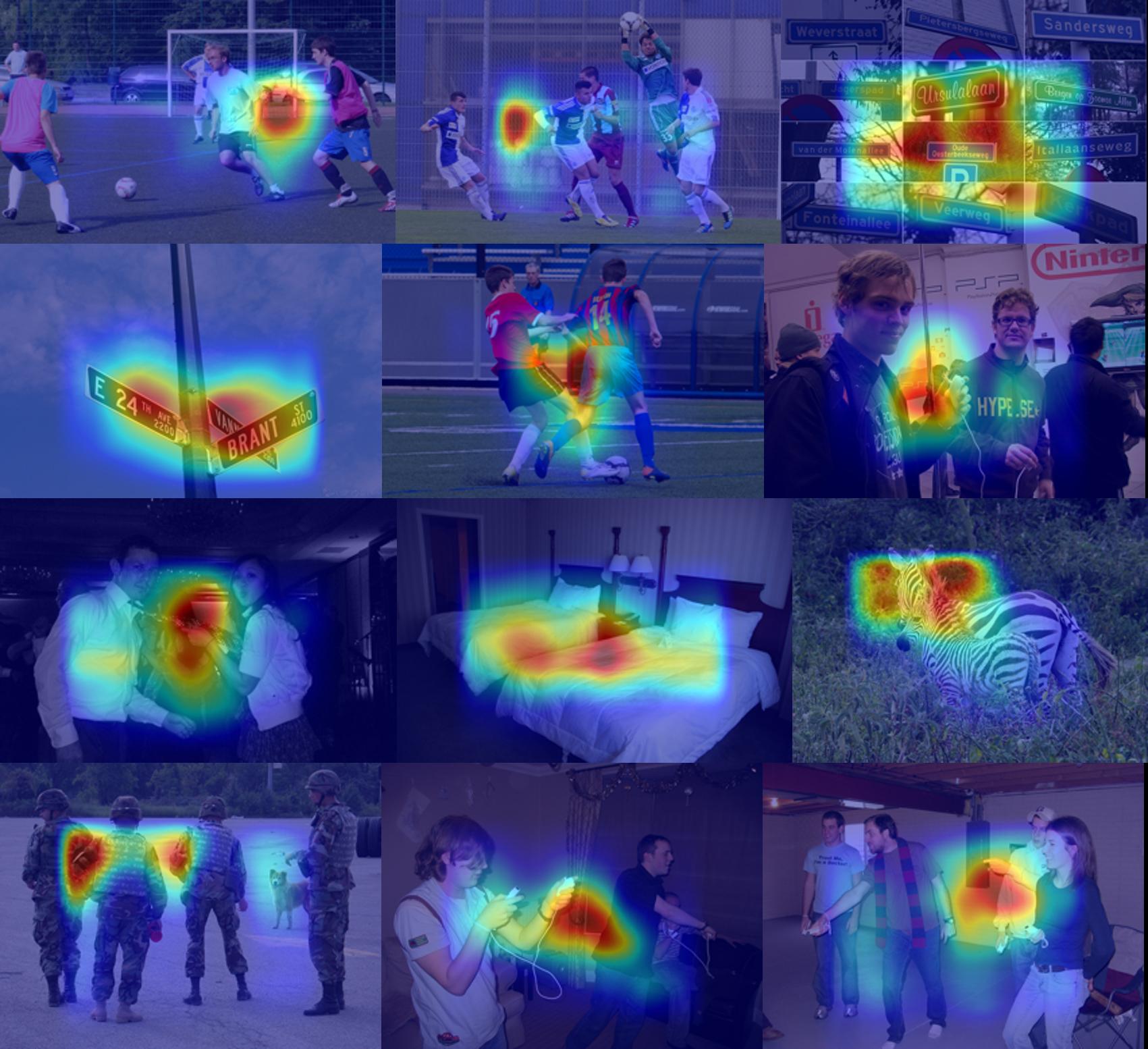}
    \caption{Multiple Instance Learning detections for \texttt{cat}, \texttt{red}, \texttt{flying} and \texttt{two} (left to right, top to bottom). View in color.}
\label{fig:mil-detect}
\end{figure}

Recently, researchers explored purely statistical approaches to guiding language models using images. Kiros et al.~\cite{kiros2013multimodal} use a log-bilinear model with bias features derived from the image to model text conditioned on the image. Also related are several contemporaneous papers \cite{mao2014explain,Vinyals2014b,Chen2014,Karpathy2014b,Donahue2014a,Xu2015,Lebret2015}. Among these, a common theme \cite{mao2014explain,Vinyals2014b,Chen2014,Karpathy2014b} has been to utilize a recurrent neural network for generating images captions by conditioning its output on image features extracted by a convolutional neural network. More recently, Donahue et al. \cite{Donahue2014a} also applied a similar model to video description. Lebret et al. \cite{Lebret2015} have investigated the use of a phrase-based model for generating captions, while Xu et al. \cite{Xu2015} have proposed a model based on visual attention.

Unlike these approaches, in this work we detect words by applying a CNN to image regions~\cite{girshickCVPR14} and integrating the information with MIL~\cite{zhangNIPS05}. We minimize {\it a priori} assumptions about how sentences should be structured by training directly from captions. Finally, in contrast to~\cite{kiros2013multimodal,mao2014explain}, we formulate the problem of generation as an optimization problem and search for the most likely sentence~\cite{ratnaparkhi2002trainable}.

\section{Word Detection}\label{sec:object_detection}

The first step in our caption generation pipeline detects a set of words that are likely to be part of the image's description. These words may belong to any part of speech, including nouns, verbs, and adjectives. We determine our vocabulary $\vocab$ using the 1000 most common words in the training captions, which cover over $92\%$ of the word occurrences in the training data (available on project webpage \footnote{\label{website}\url{http://research.microsoft.com/image_captioning}}).

\subsection{Training Word Detectors}
Given a vocabulary of words, our next goal is to detect the words from images. We cannot use standard supervised learning techniques for learning detectors, since we do not know the image bounding boxes corresponding to the words. In fact, many words relate to concepts for which bounding boxes may not be easily defined, such as \texttt{open} or \texttt{beautiful}. One possible approach is to use image classifiers that take as input the entire image. As we show in Section \ref{sec:experiments}, this leads to worse performance since many words or concepts only apply to image sub-regions. Instead, we learn our detectors using the weakly-supervised approach of Multiple Instance Learning (MIL) \cite{maron98,zhangNIPS05}.

For each word $w \in \vocab$, MIL takes as input sets of ``positive'' and ``negative'' bags of bounding boxes, where each bag corresponds to one image $i$. A bag $b_i$ is said to be positive if word $w$ is in image $i$'s description, and negative otherwise. Intuitively, MIL performs training by iteratively selecting instances within the positive bags, followed by retraining the detector using the updated positive labels.

We use a noisy-OR version of MIL \cite{zhangNIPS05}, where the probability of bag $b_i$ containing word $w$ is calculated from the probabilities of individual instances in the bag:
\begin{equation}
 1 - \prod_{j \in b_i} \left(1-p^w_{ij}\right)
\end{equation}
where $p^w_{ij}$ is the probability that a given image region $j$ in image $i$ corresponds to word $w$. We compute $p^w_{ij}$ using a multi-layered architecture \cite{krizhevskyNIPS12,simonyan14very}\footnote{We denote the CNN from \cite{krizhevskyNIPS12} as AlexNet and the 16-layer CNN from \cite{simonyan14very} as VGG for subsequent discussion. We use the code base and models available from the Caffe Model Zoo \url{https://github.com/BVLC/caffe/wiki/Model-Zoo} \cite{jia2014caffe}.}, by computing a logistic function on top of the \texttt{fc7} layer (this can be expressed as a fully connected \texttt{fc8} layer followed by a sigmoid layer):
\begin{equation}
 \frac{1}{1+\exp\left(-(\mathbf{{v}_w^t}\phi(b_{ij})+u_w)\right)},
 \label{eqn:sigmoid}
\end{equation}
where $\phi(b_{ij})$ is the \texttt{fc7} representation for image region $j$ in image $i$, and $\mathbf{v_w}$, $u_w$ are the weights and bias associated with word $w$.

We express the fully connected layers (\texttt{fc6}, \texttt{fc7}, \texttt{fc8}) of these networks as convolutions to obtain a fully convolutional network. When this fully convolutional network is run over the image, we obtain a coarse spatial response map. Each location in this response map corresponds to the response obtained by applying the original CNN to overlapping shifted regions of the input image (thereby effectively scanning different locations in the image for possible objects). We up-sample the image to make the longer side to be $565$ pixels which gives us a $12\times12$ response map at \texttt{fc8} for both \cite{krizhevskyNIPS12,simonyan14very} and corresponds to sliding a $224\times224$ bounding box in the up-sampled image with a stride of 32. The noisy-OR version of MIL is then implemented on top of this response map to generate a single probability $p^{w}_i$ for each word for each image. We use a cross entropy loss and optimize the CNN end-to-end for this task with stochastic gradient descent. We use one image in each batch and train for 3 epochs. For initialization, we use the network pre-trained on ImageNet \cite{imagenet_cvpr09}.

\subsection{Generating Word Scores for a Test Image}\label{sec:detgen}
Given a novel test image $i$, we up-sample and forward propagate the image through the CNN to obtain $p^{w}_{i}$ as described above. We do this for all words $w$ in the vocabulary $\vocab$. Note that all the word detectors have been trained independently and hence their outputs need to be calibrated. To calibrate the output of different detectors, we use the image level likelihood $p^w_i$ to compute precision on a held-out subset of the training data~\cite{hariharanCVPR2014b}. We threshold this precision value at a global threshold $\tau$, and output all words $\vocabThres$ with a precision of $\tau$ or higher along with the image level probability $p^w_{i}$, and raw score $\max_{j}{p^{w}_{ij}}$.

Figure~\ref{fig:mil-detect} shows some sample MIL detections. For each image, we visualize the spatial response map $p^{w}_{ij}$. Note that the method has not used any bounding box annotations for training, but is still able to reliably localize objects and also associate image regions with more abstract concepts.

\section{Language Generation}\label{sec:language}

\begin{table*}\centering\footnotesize
\caption{Features used in the maximum entropy language model.}
\label{tab:features}\vspace{1mm}
\begin{tabular}{@{}l@{\hspace{4mm}}c@{\hspace{4mm}}c@{\hspace{4mm}}l}
\toprule
 Feature & Type & Definition & Description\\
\midrule
 Attribute & 0/1 & $\bar{w}_{l} \in \vocabThres_{l-1}$ &
 Predicted word is in the attribute set, i.e. has been visually detected and not yet used. \\
 N-gram+& 0/1 & $\bar{w}_{l-N+1},\cdots,\bar{w}_l = \kappa$ and $\bar{w}_{l} \in \vocabThres_{l-1}$ &
 N-gram ending in predicted word is $\kappa$ and the predicted word is in the attribute set. \\
 N-gram-& 0/1 & $\bar{w}_{l-N+1},\cdots,\bar{w}_l = \kappa$ and $\bar{w}_{l} \notin \vocabThres_{l-1}$ &
 N-gram ending in predicted word is $\kappa$ and the predicted word is not in the attribute set. \\
 End & 0/1 & $\bar{w}_l = \kappa$ and $\vocabThres_{l-1} = \emptyset$ &
 The predicted word is $\kappa$ and all attributes have been mentioned. \\
 Score & $\mathbb{R}$ & score($\bar{w}_{l}$) when $\bar{w}_{l} \in \vocabThres_{l-1}$ &
 The log-probability of the predicted word when it is in the attribute set. \\
\bottomrule
\end{tabular}
\vspace{-1em}
\end{table*}

We cast the generation process as a search for the likeliest sentence conditioned on the set of visually detected words. The language model is at the heart of this process because it defines the probability distribution over word sequences. Note that despite being a statistical model, the LM can encode very meaningful information, for instance that \texttt{running} is more likely to follow \texttt{horse} than \texttt{talking}. This information can help identify false word detections and encodes a form of commonsense knowledge.

\subsection{Statistical Model}\label{ssec:lm}

To generate candidate captions for an image, we use a maximum entropy (ME) LM conditioned on the set of visually detected words. The ME LM estimates the probability of a word $w_l$ conditioned on the preceding words $w_1, w_2, \cdots, w_{l-1}$, as well as the set of words with high likelihood detections $\vocabThres_l \subset \vocabThres$ that have yet to be mentioned in the sentence. The motivation of conditioning on the unused words is to encourage all the words to be used, while avoiding repetitions. The top 15 most frequent closed-class words\footnote{The top 15 frequent closed-class words are \texttt{a}, \texttt{on}, \texttt{of}, \texttt{the}, \texttt{in}, \texttt{with}, \texttt{and}, \texttt{is}, \texttt{to}, \texttt{an}, \texttt{at}, \texttt{are}, \texttt{next}, \texttt{that} and \texttt{it}.} are removed from the set $\vocabThres$ since they are detected in nearly every image (and are trivially generated by the LM). It should be noted that the detected words are usually somewhat noisy. Thus, when the end of sentence token is being predicted, the set of remaining words may still contain some words with a high confidence of detection.

Following the definition of an ME LM \cite{Berger1996}, the word probability conditioned on preceding words and remaining objects can be written as:\vspace{-.5em}

{\scriptsize
\begin{align}
	&\Pr(w_l = \bar{w}_l | \bar{w}_{l-1}, \cdots, \bar{w}_1, \sosTok,
	\vocabThres_{l-1}) =\nonumber \\
	& \frac{\exp \left[
		\sum_{k=1}^K \lambda_k f_k(\bar{w}_l, \bar{w}_{l-1}, \cdots, \bar{w}_1, \sosTok,
		\vocabThres_{l-1})
	\right]}
	{\sum_{v \in \mathcal{V} \cup \eosTok} \exp \left[
		\sum_{k=1}^K \lambda_k f_k(v, \bar{w}_{l-1}, \cdots, \bar{w}_1, \sosTok,
		\vocabThres_{l-1})
	\right]}
	\label{eq:melm}
\end{align}}where $\sosTok$ denotes the start-of-sentence token, $\bar{w}_j \in \mathcal{V} ~ \cup \eosTok$, and $f_k(w_l, \cdots, w_1, \vocabThres_{l-1})$ and $\lambda_k$ respectively denote the $k$-th max-entropy feature and its weight. The basic discrete ME features we use are summarized in Table \ref{tab:features}. These features form our ``baseline'' system. It has proven effective to extend this with a ``score'' feature, which evaluates to the log-likelihood of a word according to the corresponding visual detector. We have also experimented with distant bigram features \cite{lau1993trigger} and continuous space log-bilinear features \cite{Mnih2007,Mnih2012}, but while these improved PPLX significantly, they did not improve BLEU, METEOR or human preference, and space restrictions preclude further discussion.

To train the ME LM, the objective function is the log-likelihood of the captions conditioned on the corresponding set of detected objects, i.e.:

\vspace{-2mm}{\small
\begin{align}
	L(\Lambda)  = \sum_{s=1}^S \sum_{l=1}^{\#{(s)}}
	\log \Pr(\bar{w}_l^{(s)} | \bar{w}_{l-1}^{(s)}, \cdots, \bar{w}_1^{(s)}, \sosTok,
	\vocabThres_{l-1}^{(s)})
\end{align}}%
where the superscript $(s)$ denotes the index of sentences in the training data, and $\#{(s)}$ denotes the length of the sentence. The noise contrastive estimation (NCE) technique is used to accelerate the training by avoiding the calculation of the exact denominator in (\ref{eq:melm}) \cite{Mnih2012}. In the generation process, we use the unnormalized NCE likelihood estimates, which are far more efficient than the exact likelihoods, and produce very similar outputs. However, all PPLX numbers we report are computed with exhaustive normalization. The ME features are implemented in a hash table as in \cite{mikolov2011strategies}. In our experiments, we use N-gram features up to 4-gram and 15 contrastive samples in NCE training.

\subsection{Generation Process}

During generation, we perform a left-to-right beam search similar to the one used in \cite{Ratnaparkhi2000}. This maintains a stack of length $l$ partial hypotheses. At each step in the search, every path on the stack is extended with a set of likely words, and the resulting length $l+1$ paths are stored. The top $k$ length $l+1$ paths are retained and the others pruned away.

We define the possible extensions to be the end of sentence token $\eosTok$, the 100 most frequent words, the set of attribute words that remain to be mentioned, and all the words in the training data that have been observed to follow the last word in the hypothesis.  Pruning is based on the likelihood of the partial path. When $\eosTok$ is generated, the full path to $\eosTok$ is removed from the stack and set aside as a completed sentence. The process continues until a maximum sentence length $L$ is reached.

After obtaining the set of completed sentences $\mathcal{C}$, we form an $M$-best list as follows. Given a target number of $T$ image attributes to be mentioned, the sequences in $\mathcal{C}$ covering at least $T$ objects are added to the $M$-best list, sorted in descending order by the log-likelihood. If there are less than $M$ sequences covering at least $T$ objects found in $\mathcal{C}$, we reduce $T$ by 1 until $M$ sequences are found.

\section{Sentence Re-Ranking}\label{sec:rerank}

Our LM produces an $M$-best set of sentences. Our final stage uses MERT \cite{Och2003} to re-rank the $M$ sentences. MERT uses a linear combination of features computed over an entire sentence, shown in Table \ref{tab:MERT}. The MERT model is trained on the $M$-best lists for the validation set using the BLEU metric, and applied to the $M$-best lists for the test set. Finally, the best sequence after the re-ranking is selected as the caption of the image.  Along with standard MERT features, we introduce a new multimodal semantic similarity model, discussed below.  

\begin{table}\centering\small
\caption{Features used by MERT.}
\vspace{1mm}\label{tab:MERT}
\begin{tabular}{l}
\toprule
	1. The log-likelihood of the sequence.\\
	2. The length of the sequence.\\
	3. The log-probability per word of the sequence.\\
	4. The logarithm of the sequence's rank in the log-likelihood.\\
	5. 11 binary features indicating whether the number\\
	   $\quad$ of mentioned objects is $x$ ($x=0,\ldots,10)$.\\
	6. The DMSM score between the sequence and the image.\\
\bottomrule
\end{tabular}
\vspace{-1.5em}
\end{table}

\subsection {Deep Multimodal Similarity Model}

To model global similarity between images and text, we develop a Deep Multimodal Similarity Model (DMSM). The DMSM learns two neural networks that map images and text fragments to a common vector representation. We measure similarity between images and text by measuring cosine similarity between their corresponding vectors. This cosine similarity score is used by MERT to re-rank the sentences. The DMSM is closely related to the unimodal Deep Structured Semantic Model (DSSM) \cite{huang2013,Shen2014}, but extends it to the multimodal setting. The DSSM was initially proposed to model the semantic relevance between textual search queries and documents, and is extended in this work to replace the query vector in the original DSSM by the image vector computed from the deep convolutional network.

The DMSM consists of a pair of neural networks, one for mapping each input modality to a common semantic space, which are trained jointly. In training, the data consists of a set of image/caption pairs. The loss function minimized during training represents the negative log posterior probability of the caption given the corresponding image.

\textbf{Image model}: We map images to semantic vectors using the same CNN (AlexNet / VGG) as used for detecting words in Section \ref{sec:object_detection}. We first finetune the networks on the COCO dataset for the full image classification task of predicting the words occurring in the image caption. We then extract out the \texttt{fc7} representation from the finetuned network and stack three additional fully connected layers with \textit{tanh} non-linearities on top of this representation to obtain a final representation of the same size as the last layer of the text model. We learn the parameters in these additional fully connected layers during DMSM training.

\textbf{Text model}: The text part of the DMSM maps text fragments to semantic vectors, in the same manner as in the original DSSM. In general, the text fragments can be a full caption. Following \cite{huang2013} we convert each word in the caption to a letter-trigram count vector, which uses the count distribution of context-dependent letters to represent a word. This representation has the advantage of reducing the size of the input layer while generalizing well to infrequent, unseen and incorrectly spelled words. Then following \cite{Shen2014}, this representation is forward propagated through a deep convolutional neural network to produce the semantic vector at the last layer.

\textbf{Objective and training}: We define the relevance $R$ as the cosine similarity between an image or query ($Q$) and a text fragment or document ($D$) based on their representations $y_Q$ and $y_D$ obtained using the image and text models:
{\small $R(Q, D) = \text{cosine}(y_Q, y_D) = ({y_Q}^Ty_D) / \Vert y_Q \Vert\Vert y_D \Vert$}.
For a given image-text pair, we can compute the posterior probability of the text being relevant to the image via:\vspace{-.5em}

{\small\begin{equation}
P(D | Q) = \frac{\exp(\gamma R(Q,D))}{\Sigma_{D' \in \mathbb{D}}\exp(\gamma R(Q, D'))}\end{equation}}\vspace{-.5em}

Here $\gamma$ is a smoothing factor determined using the validation set, which is 10 in our experiments. $\mathbb{D}$ denotes the set of all candidate documents (captions) which should be compared to the query (image). We found that restricting $\mathbb{D}$ to one matching document $D^+$ and a fixed number $N$ of randomly selected non-matching documents $D^-$ worked reasonably well, although using noise-contrastive estimation could further improve results. Thus, for each image we select one relevant text fragment and $N$ non-relevant fragments to compute the posterior probability. $N$ is set to 50 in our experiments. During training, we adjust the model parameters $\Lambda$ to minimize the negative log posterior probability that the relevant captions are matched to the images:\vspace{-.5em}

{\small\begin{equation}\label{eq:dmsm_objective}
L(\Lambda) = -\log\prod_{(Q, D^{+})}P(D^+|Q)
\end{equation}}\vspace{-.75em}

\section{Experimental Results}\label{sec:experiments}

\begin{table*}\centering\small
\caption{Average precision (AP) and Precision at Human Recall (PHR) \cite{capeval2015} for words with different parts of speech (NN: Nouns, VB: Verbs, JJ: Adjectives, DT: Determiners, PRP: Pronouns, IN: Prepositions). Results are shown using a chance classifier, full image classification, and Noisy OR multiple instance learning with AlexNet \cite{krizhevskyNIPS12} and VGG \cite{simonyan14very} CNNs.}
\label{tab:ap}\vspace{2mm}
\scalebox{0.8}{
\begin{tabular}{llccccccccccccccccc}
                       & \multicolumn{8}{c}{Average Precision} && \multicolumn{8}{c}{Precision at Human Recall} \\ \cmidrule(r){2-9} \cmidrule(r){11-18}
                          & NN   & VB   & JJ   & DT   & PRP  & IN   & Others & All  && NN   & VB   & JJ   & DT   & PRP  & IN   & Others & All \\ \midrule
 Count                    & 616  & 176  & 119  & 10   & 11   & 38   & 30     & 1000 &&      &      &      &      &      &      &        & \\ \midrule
 Chance                   & 2.0  & 2.3  & 2.5  & 23.6 & 4.7  & 11.9 & 7.7    & 2.9  &&      &      &      &      &      &      &        & \\
 Classification (AlexNet) & 32.4 & 16.7 & 20.7 & 31.6 & 16.8 & 21.4 & 15.6   & 27.1 && 39.0 & 27.7 & 37.0 & 37.3 & 26.2 & 31.5 & 25.0   & 35.9 \\
 Classification (VGG)     & 37.0 & 19.4 & 22.5 & 32.9 & 19.4 & 22.5 & 16.9   & 30.8 && 45.3 & 31.0 & 37.1 & 40.2 & 29.6 & 33.9 & 25.5   & 40.6 \\
 MIL            (AlexNet) & 36.9 & 18.0 & 22.9 & 31.7 & 16.8 & 21.4 & 15.2   & 30.4 && 46.0 & 29.4 & 40.1 & 37.9 & 25.9 & 31.5 & 21.6   & 40.8 \\
 MIL            (VGG)     & 41.4 & 20.7 & 24.9 & 32.4 & 19.1 & 22.8 & 16.3   & 34.0 && 51.6 & 33.3 & 44.3 & 39.2 & 29.4 & 34.3 & 23.9   & 45.7 \\
 Human Agreement          &      &      &      &      &      &      &        &      && 63.8 & 35.0 & 35.9 & 43.1 & 32.5 & 34.3 & 31.6   & 52.8 \\
\bottomrule
\end{tabular}}
\end{table*}

\begin{figure*}
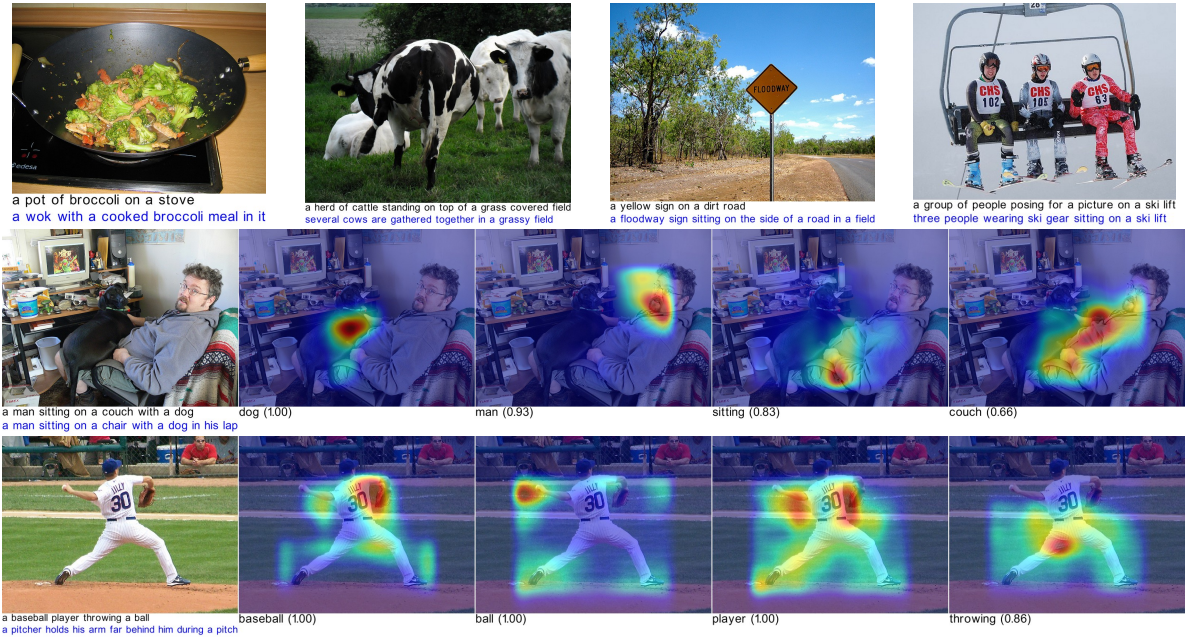

\centering
 \insertH{0.17}{{coco-vis/v3/582015-test.all-341084}.jpg} $\quad$
 \insertH{0.17}{{coco-vis/v3/725998-test.all-225364}.jpg} $\quad$
 \insertH{0.17}{{coco-vis/v3/691878-test.all-177607}.jpg} $\quad$
 \insertH{0.17}{{coco-vis/v3/559717-test.all-356511}.jpg} \\
 \insertW{.9}{{coco-vis/v4/295483-test.all-285789}.jpg}
 \insertW{.9}{{coco-vis/v4/923025-test.all-326916}.jpg}
 \caption{Qualitative results for several randomly chosen images on the Microsoft COCO dataset, with our generated caption (black) and a human caption (blue) for each image. In the bottom two rows we show localizations for the words used in the sentences. More examples can be found on the project website$^{\ref{website}}$.}
\label{fig:qual}
\vspace{-1em}
\end{figure*}

We next describe the datasets used for testing, followed by an evaluation of our approach for word detection and experimental results on sentence generation.

\subsection{Datasets}

Most of our results are reported on the Microsoft COCO dataset \cite{linECCV14,capeval2015}. The dataset contains 82,783 training images and 40,504 validation images. The images create a challenging testbed for image captioning since most images contain multiple objects and significant contextual information. The COCO dataset provides 5 human-annotated captions per image. The test annotations are not available, so we split the validation set into validation and test sets\footnote{We split the COCO train/val set ito 82,729 train/20243 val/20244 test. Unless otherwise noted, test results are reported on the 20444 images from the validation set.}.

For experimental comparison with prior papers, we also report results on the PASCAL sentence dataset \cite{Rashtchian2010}, which contains 1000 images from the 2008 VOC Challenge \cite{PASCAL}, with 5 human captions each.

\subsection{Word Detection}

To gain insight into our weakly-supervised approach for word detection using MIL, we measure its accuracy on the word classification task: If a word is used in at least one ground truth caption, it is included as a positive instance. Note that this is a challenging task, since conceptually similar words are classified separately; for example, the words cat/cats/kitten, or run/ran/running all correspond to different classes. Attempts at adding further supervision, e.g., in the form of lemmas, did not result in significant gains.

Average Precision (AP) and Precision at Human Recall (PHR) \cite{capeval2015} results for different parts of speech are shown in Table \ref{tab:ap}. We report two baselines. The first (Chance) is the result of randomly classifying each word. The second (Classification) is the result of a whole image classifier which uses features from AlexNet or VGG CNN \cite{krizhevskyNIPS12,simonyan14very}. These features were fine-tuned for this word classification task using a logistic regression loss.

As shown in Table \ref{tab:ap}, the MIL NOR approach improves over both baselines for all parts of speech, demonstrating that better localization can help predict words. In fact, we observe the largest improvement for nouns and adjectives, which often correspond to concrete objects in an image sub-region.  Results for both classification and MIL NOR are lower for parts of speech that may be less visually informative and difficult to detect, such as adjectives (e.g., \texttt{few}, which has an AP of 2.5), pronouns (e.g., \texttt{himself}, with an AP of 5.7), and prepositions (e.g., \texttt{before}, with an AP of 1.0). In comparison words with high AP scores are typically either visually informative (\texttt{red}: AP 66.4, \texttt{her}: AP 45.6) or associated with specific objects (\texttt{polar}: AP 94.6, \texttt{stuffed}: AP 74.2). Qualitative results demonstrating word localization are shown in Figures \ref{fig:mil-detect} and \ref{fig:qual}.

\begin{figure}\centering
 \insertH{0.145}{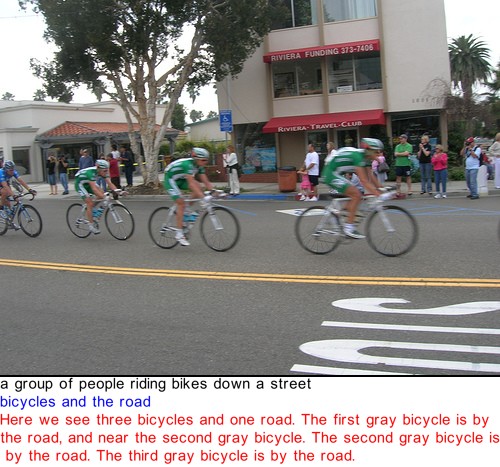}
 \insertH{0.145}{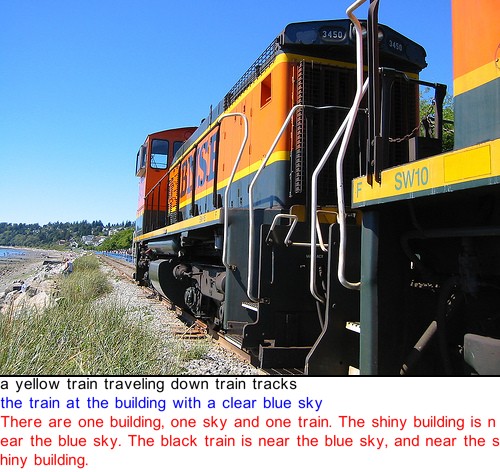}
 \insertH{0.145}{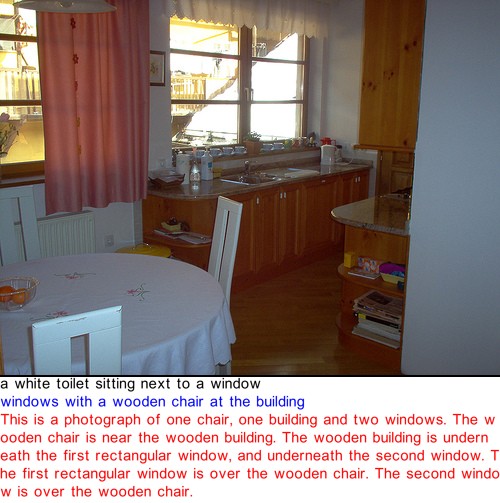}
\caption{Qualitative results for images on the PASCAL sentence dataset. Captions using our approach (black), Midge \cite{mitchell2012midge} (blue) and Baby Talk \cite{kulkarni2011baby} (red) are shown.}
\label{fig:PASCAL}
\vspace{-1em}
\end{figure}

\subsection{Caption Generation}\label{sec:caption_generation}
\begin{table*}\centering\small
\caption{Caption generation performance for seven variants of our system on the Microsoft COCO dataset.   We report performance on our held out test set (half of the validation set). We report Perplexity (PPLX), BLEU and METEOR, using 4 randomly selected caption references. Results from human studies of subjective performance are also shown, with error bars in parentheses. Our final System ``VGG+Score+DMSM+ft'' is ``same or better'' than human 34\% of the time.}
\label{tab:mainresult}\vspace{1.5mm}
\begin{tabular}{lcccccccc}
\toprule
System & {\sc pplx} & {\sc bleu} & {\sc meteor} 
 & $\approx$human & $>$human & $\ge$human \\
\midrule
1. Unconditioned                     & 24.1 & 1.2\%  &  6.8\%  \\
2. Shuffled Human                    & --   & 1.7\%  &  7.3\%  \\
3. Baseline                          & 20.9 & 16.9\% & 18.9\% &  9.9\% ($\pm$1.5\%) & 2.4\% ($\pm$0.8\%) & 12.3\% ($\pm$1.6\%) \\
4. Baseline+Score                    & 20.2 & 20.1\% & 20.5\% & 16.9\% ($\pm$2.0\%) & 3.9\% ($\pm$1.0\%) & 20.8\% ($\pm$2.2\%) \\
5. Baseline+Score+{\small DMSM}      & 20.2 & 21.1\% & 20.7\% & 18.7\% ($\pm$2.1\%) & 4.6\% ($\pm$1.1\%) & 23.3\% ($\pm$2.3\%) \\
6. Baseline+Score+{\small DMSM}+ft   & 19.2 & 23.3\% & 22.2\% & -- & -- & --\\
7. VGG+Score+ft                      & 18.1 & 23.6\% & 22.8\% & -- & -- & --\\
8. VGG+Score+{\small DMSM}+ft        & 18.1 & 25.7\% & 23.6\% & 26.2\% ($\pm$2.1\%)	& 7.8\% ($\pm$1.3\%) & {\bf 34.0\%} ($\pm$2.5\%)\\
\midrule
Human-written captions               & --   & 19.3\% & 24.1\% \\
\bottomrule
\end{tabular}
\vspace{-1em}
\end{table*}

We next describe our caption generation results, beginning with a short discussion of evaluation metrics.

\textbf{Metrics:} The sentence generation process is measured using both automatic metrics and human studies. We use three different automatic metrics: PPLX, BLEU \cite{papineni2002bleu}, and METEOR \cite{banerjee2005meteor}. PPLX (perplexity) measures the uncertainty of the language model, corresponding to how many bits on average would be needed to encode each word given the language model. A lower PPLX indicates a better score. BLEU \cite{papineni2002bleu} is widely used in machine translation and measures the fraction of N-grams (up to 4-gram) that are in common between a hypothesis and a reference or set of references; here we compare against 4 randomly selected references. METEOR \cite{banerjee2005meteor} measures unigram precision and recall, extending exact word matches to include similar words based on WordNet synonyms and stemmed tokens.  We additionally report performance on the metrics made available from the MSCOCO captioning challenge,\footnote{\url{http://mscoco.org/dataset/\#cap2015}} which includes scores for BLEU-1 through BLEU-4, METEOR, CIDEr \cite{VedantamCORR14}, and ROUGE-L \cite{LinACL04}. 

All of these automatic metrics are known to only roughly correlate with human judgment \cite{elliot2014}. We therefore include human evaluation to further explore the quality of our models. Each task presents a human (Mechanical Turk worker) with an image and two captions: one is automatically generated, and the other is a human caption. The human is asked to select which caption better describes the image, or to choose a ``same'' option when they are of equal quality.  In each experiment, 250 humans were asked to compare 20 caption pairs each, and 5 humans judged each caption pair. We used Crowdflower, which automatically filters out spammers. The ordering of the captions was randomized to avoid bias, and we included four check-cases where the answer was known and obvious; workers who missed any of these were excluded. The final judgment is the majority vote of the judgment of the 5 humans. In ties, one-half of a count is distributed to the two best answers.  We also compute errors bars on the human results by taking 1000 bootstrap resamples of the majority vote outcome (with ties), then reporting the difference between the mean and the 5th or 95th percentile (whichever is farther from the mean). 

\textbf{Generation results:} Table \ref{tab:mainresult} summarizes our  results on the Microsoft COCO dataset. We provide several baselines for experimental comparison, including two baselines that measure the complexity of the dataset: Unconditioned, which generates sentences by sampling an $N$-gram LM without knowledge of the visual word detectors; and Shuffled Human, which randomly picks another human generated caption from another image. Both the BLEU and METEOR scores are very low for these approaches, demonstrating the variation and complexity of the Microsoft COCO dataset.

We provide results on seven variants of our end-to-end approach: Baseline is based on visual features from AlexNet and uses the ME LM with all the discrete features as described in Table~\ref{tab:features}. Baseline+Score adds the feature for the word detector score into the ME LM. Both of these versions use the same set of sentence features (excluding the DMSM score) described in Section~\ref{sec:rerank} when re-ranking the captions using MERT. Baseline+Score+DMSM uses the same ME LM as Baseline+Score, but adds the DMSM score as a feature for re-ranking. Baseline+Score+DMSM+ft adds finetuning. VGG+Score+ft and VGG+Score+DMSM+ft are analogous to Baseline+Score and Baseline+Score+DMSM but use finetuned VGG features. Note: the AlexNet baselines without finetuning are from an early version of our system which used object proposals from \cite{zitnickECCV14} instead of dense scanning.

As shown in Table \ref{tab:mainresult}, the PPLX of the ME LM with and without the word detector score feature is roughly the same. But, BLEU and METEOR improve with addition of the word detector scores in the ME LM. Performance improves further with addition of the DMSM scores in re-ranking. Surprisingly, the BLEU scores are actually above those produced by human generated captions (25.69\% vs.~19.32\%). Improvements in performance using the DMSM scores with the VGG model are statistically significant as measured by 4-gram overlap and METEOR per-image (Wilcoxon signed-rank test, p $<$ .001).

We also evaluated an approach (not shown) with whole-image classification rather than MIL.  We found this approach to under-perform relative to MIL in the same setting (for example, using the VGG+Score+DMSM+ft setting, PPLX=18.9, BLEU=21.9\%, METEOR=21.4\%). This suggests that integrating information about words associated to image regions with MIL leads to improved performance over image classification alone.

The VGG+Score+DMSM approach produces captions that are judged to be of the same or better quality than human-written descriptions 34\% of the time, which is a significant improvement over the Baseline results. Qualitative results are shown in Figure \ref{fig:qual}, and many more are available on the project website.

\begin{table}
\caption{Official COCO evaluation server results on test set (40,775 images). First row show results using 5 reference captions, second row, 40 references. Human results reported in parentheses.\vspace{1mm}}
\centering
\resizebox{\columnwidth}{!} {
\begin{tabular}{@{}ccccccccccccccc@{}}
 \toprule
 & {\sc cide}r & {\sc bleu-{\small 4}} & {\sc bleu-{\small 1}} & {\sc rouge-l} & {\sc meteor} \\
 \midrule
 {\bf [5]} & .912 (.854) & .291 (.217) & .695 (.663) &  .519 (.484) & .247 (.252)\\
 {\bf [40]} & .925 (.910) & .567 (.471) & .880 (.880) & .662 (.626) & .331 (.335) \\
 \bottomrule
\end{tabular}}
\label{tab:official_coco}
\vspace{-1em}
\end{table}

\textbf{COCO evaluation server results:} We further generated the captions for the images in the actual COCO test set consisting of 40,775 images (human captions for these images are not available publicly), and evaluated them on the COCO evaluation server. These results are summarized in Table \ref{tab:official_coco}. Our system gives a BLEU-4 score of 29.1\%, and equals or surpasses human performance on 12 of the 14 metrics reported -- the only system to do so. These results are also state-of-the-art on all 14 reported metrics among the four other results available publicly at the time of writing this paper. In particular, our system is the only one exceeding human CIDEr scores, which has been specifically proposed for evaluating image captioning systems \cite{VedantamCORR14}.

To enable direct comparison with previous work on automatic captioning, we also test on the PASCAL sentence dataset \cite{Rashtchian2010}, using the 847 images tested for both the Midge \cite{mitchell2012midge} and Baby Talk \cite{kulkarni2011baby} systems. We show significantly improved results over the Midge \cite{mitchell2012midge} system, as measured by both BLEU and METEOR (2.0\%~vs.~17.6\% BLEU and 9.2\%~vs.~19.2\% METEOR).\footnote{Baby Talk generates long, multi-sentence captions, making comparison by BLEU/METEOR difficult; we thus exclude evaluation here.} To give a basic sense of the progress quickly being made in this field, Figure \ref{fig:PASCAL} shows output from the system on the same images.\footnote{Images were selected visually, without viewing system captions.}

\section{Conclusion}\label{sec:conclusion}

This paper presents a new system for generating novel captions from images. The system trains on images and corresponding captions, and learns to extract nouns, verbs, and adjectives from regions in the image. These detected words then guide a language model to generate text that reads well and includes the detected words. Finally, we use a global deep multimodal similarity model introduced in this paper to re-rank candidate captions

At the time of writing, our system is state-of-the-art on all 14 official metrics of the COCO image captioning task, and equal to or exceeding human performance on 12 out of the 14 official metrics. Our generated captions have been judged by humans (Mechanical Turk workers) to be equal to or better than human-written captions 34\% of the time.

{\small\bibliographystyle{ieee}\bibliography{biblio}}

\end{document}